\documentclass[10pt,twocolumn,letterpaper]{article}

\usepackage[pagenumbers]{cvpr} % To force page numbers, e.g. for an arXiv version
\usepackage{tabularx}
\usepackage{pifont}
\usepackage{marvosym}
\usepackage{booktabs}
\usepackage{multirow}
\usepackage{makecell}
\usepackage{adjustbox}
\usepackage{colortbl}

\definecolor{cvprblue}{rgb}{0.21,0.49,0.74}
\usepackage[pagebackref,breaklinks,colorlinks,allcolors=cvprblue]{hyperref}

\def\paperID{*****} % *** Enter the Paper ID here
\def\confName{CVPR}
\def\confYear{2026}

\title{HEDGE: Heterogeneous Ensemble for Detection of AI-GEnerated Images in the Wild}

\author{ 
\small Fei Wu$^{1}$, Dagong Lu$^{2}$, Mufeng Yao$^{2}$, Xinlei Xu$^{2}$,  Fengjun Guo$^{2,\text{\Letter}}$ 
\\
$^1$ \small Shanghai Jiao Tong University \\
$^2$ \small INTSIG Information \\
}

\begin{document}
\maketitle
\let\thefootnote\relax\footnotetext{$^{\text{\Letter}}$ Corresponding author.}
\begin{abstract}

Robust detection of AI-generated images in the wild remains challenging due to the rapid evolution of generative models and varied real-world distortions.
We argue that relying on a single training regime, resolution, or backbone is insufficient to handle all conditions, and that structured heterogeneity across these dimensions is essential for robust detection.
To this end, we propose \textbf{HEDGE}, a \textbf{H}eterogeneous \textbf{E}nsemble for \textbf{D}etection of AI-\textbf{GE}nerated images, that introduces complementary detection routes along three axes: diverse training data with strong augmentation, multi-scale feature extraction, and backbone heterogeneity.
Specifically, Route~A progressively constructs DINOv3-based detectors through staged data expansion and augmentation escalation, Route~B incorporates a higher-resolution branch for fine-grained forensic cues, and Route~C adds a MetaCLIP2-based branch for backbone diversity.
All outputs are fused via logit-space weighted averaging, refined by a lightweight dual-gating mechanism that handles branch-level outliers and majority-dominated fusion errors.
HEDGE achieves 4th place in the NTIRE 2026 Robust AI-Generated Image Detection in the Wild Challenge and attains state-of-the-art performance with strong robustness on multiple AIGC image detection benchmarks.

\end{abstract}
\section{Introduction}
\label{sec:intro}

Recent advances in generative models---from GANs and diffusion models to autoregressive architectures---have made AI-generated images increasingly photorealistic, diverse, and easily accessible~\cite{Rombach_2022_CVPR,wu2025qwenimagetechnicalreport}.
This rapid progress has raised growing concerns about reliable AI-generated image detection in high-stakes scenarios such as social media moderation and forensic investigation~\cite{roettgers2021tiktok,cnn2024deepfake}.
However, the core challenge is no longer closed-set discrimination on limited generators, but robust generalization under continuously evolving generation pipelines and uncontrolled real-world distortions.

Existing studies reveal several persistent bottlenecks.
First, methods built on frequency-domain or other low-level artifacts are vulnerable to common post-processing such as resizing and JPEG compression, which attenuate the spectral signatures these detectors rely on~\cite{tan2024rethinking,zhong2023patchcraft,wang2023dire,tan2024frequency,yan2026dual}.
Second, approaches that leverage pre-trained feature extraction models tend to overfit the training distribution and learn superficial distributional shortcuts---such as biases in semantics, resolution, or data format between real and generated images---rather than genuine forgery traces, undermining generalization to unseen generators and real-world conditions~\cite{ojha2023towards,chen2025dual,yang2025d,yan2025a}.
Moreover, recent MLLM-based explainable approaches, while providing richer reasoning outputs, generally suffer from limited detection robustness and high computational costs~\cite{xu2024fakeshield,kang2025legion,ji2026fakexplain}.

Our approach builds on the observation that modern vision foundation models (VFMs) already encode strong forensic priors.
Recent studies show that self-supervised VFMs such as DINOv3~\cite{simeoni2025dinov3} outperform purpose-built detectors by substantial margins on in-the-wild benchmarks, even with a simple linear classifier~\cite{zhou2025brought}, while vision-language models such as MetaCLIP2~\cite{chuang2025meta} provide complementary representations through contrastive image-text alignment.
However, no single training schedule, resolution, or backbone dominates across all conditions, and diverse representations can implicitly cover a wider range of distribution scenarios.
This motivates a shift from seeking a single stronger model to combining heterogeneous detection routes in a structured manner.

Based on this principle, we propose a \textbf{three-route heterogeneous framework} that introduces structured diversity along three axes: \textit{diverse training data with strong augmentation}, \textit{multi-scale feature extraction}, and \textit{backbone heterogeneity}.
Route~A progressively constructs DINOv3-Huge detectors through continuation training with diversified data sources and augmentation escalation, building a family of detectors with varied robustness profiles.
Route~B retains the same DINOv3 backbone but operates at a higher input resolution, fusing multi-scale forensic cues to capture fine-grained forgery traces that are suppressed at standard scales.
Route~C replaces the backbone with MetaCLIP2-Giant to break the homogeneity of the DINOv3-based routes, mitigating the risk of correlated errors introduced by relying on a single backbone family.
Their predictions are fused in the logit space, followed by a dual-gating robust decision module that suppresses branch-level outliers and corrects inconsistent fusion under strong cross-route consensus.

Our main contributions are summarized as follows:
\begin{itemize}
	\item We propose a three-route heterogeneous framework that diversifies training data and augmentation, input resolution, and backbone architecture, achieving complementary coverage beyond any single detector.
	\item We design a lightweight dual-gating mechanism over logit-space fusion that suppresses branch-level outliers and corrects majority-dominated errors via cross-route consensus.
	\item Our method achieves 4th place in the NTIRE 2026 Robust AI-Generated Image Detection in the Wild Challenge~\cite{ntire26aigendet} and state-of-the-art results on multiple public benchmarks.
\end{itemize}

\section{Related Work}
\label{sec:related}

\subsection{Generalizable AIGC Image Detection}

Early AIGC detectors relied on generator-specific artifacts or dataset-specific shortcuts, limiting their transferability to unseen generators~\cite{ojha2023towards,tan2024rethinking,zhong2023patchcraft}.
Subsequent work improves generalization along multiple axes:
data alignment methods such as DDA~\cite{chen2025dual} and B-Free~\cite{guillaro2025bias} reduce spurious biases between real and fake distributions;
multi-source training frameworks like D$^3$~\cite{yang2025d} and GAPL~\cite{qin2025Scaling} leverage diverse generator sources to learn shared forgery cues and improve cross-generator generalization;
and fine-grained approaches including PPL~\cite{yang2025all} and HiDA-Net~\cite{mu2026no} exploit multi-patch and resolution-aware cues for distortion robustness.
Despite steady progress, benchmark evaluations on Chameleon~\cite{yan2025a}, AIGIBench~\cite{li2025artificial}, and RealChain~\cite{liu2025beyond} consistently show sharp performance degradation under realistic post-processing and chain degradation, indicating that purpose-built detectors still fall short when deployment conditions shift along multiple dimensions simultaneously.

\subsection{VFM-Based Image Forensics}

A recent study~\cite{zhou2025brought} offers a different perspective: modern vision foundation models (VFMs) already encode strong forensic priors.
Even a simple linear classifier on a frozen DINOv3 achieves state-of-the-art detection performance across multiple in-the-wild benchmarks, demonstrating that self-supervised VFMs encode rich forensic representations without task-specific design.
Building on these findings, REM~\cite{liu2025beyond} uses frozen DINOv3 as a semantic anchor for real-distribution envelope estimation, and MIRROR~\cite{liu2026mirror} projects DINOv3 patch features onto a learned real-image manifold for reference--comparison detection.
These results confirm VFM representations as a strong forensic foundation; however, a single pre-trained feature space with a simple classifier shows limited fitting capacity as the deployment distribution grows complex~\cite{yang2025d}, and different backbone families---self-supervised models vs.\ vision-language models---are likely to offer orthogonal strengths due to their distinct pre-training objectives.
Our work builds on these VFM-based findings and addresses this limitation through structured heterogeneity: combining multiple backbones, resolutions, and training regimes to exploit their complementary forensic strengths.

\subsection{MLLM-Based Image Forensics}

Recent multimodal large language models (MLLMs) have been applied to image forensics to enhance interpretability.
Representative efforts generate textual rationales alongside detection outputs~\cite{xu2024fakeshield,kang2025legion,ji2026fakexplain}, employ structured reasoning or multi-agent debate for forgery analysis~\cite{tan2025veritas,liang2025evidence,huang2025unishield}, or perform iterative zoom-in analysis to localize tampered regions~\cite{ji2025zoom}.
However, despite richer explanatory outputs, these methods generally exhibit weaker image-level detection robustness: strong language reasoning does not automatically translate into stable out-of-distribution detection, and MLLMs tend to rely on hallucination-prone semantic cues rather than genuine visual forensic evidence~\cite{xiao2026unveiling,wang2025forensics}.
Our work takes a different angle: we focus on maximizing image-level detection accuracy and robustness through heterogeneous visual routes and consensus-aware fusion, which can in turn provide a more reliable foundation for downstream explainable forensic systems.

\section{Method}
\label{sec:method}

\begin{figure*}[!t]
    \centering
    \includegraphics[width=\linewidth]{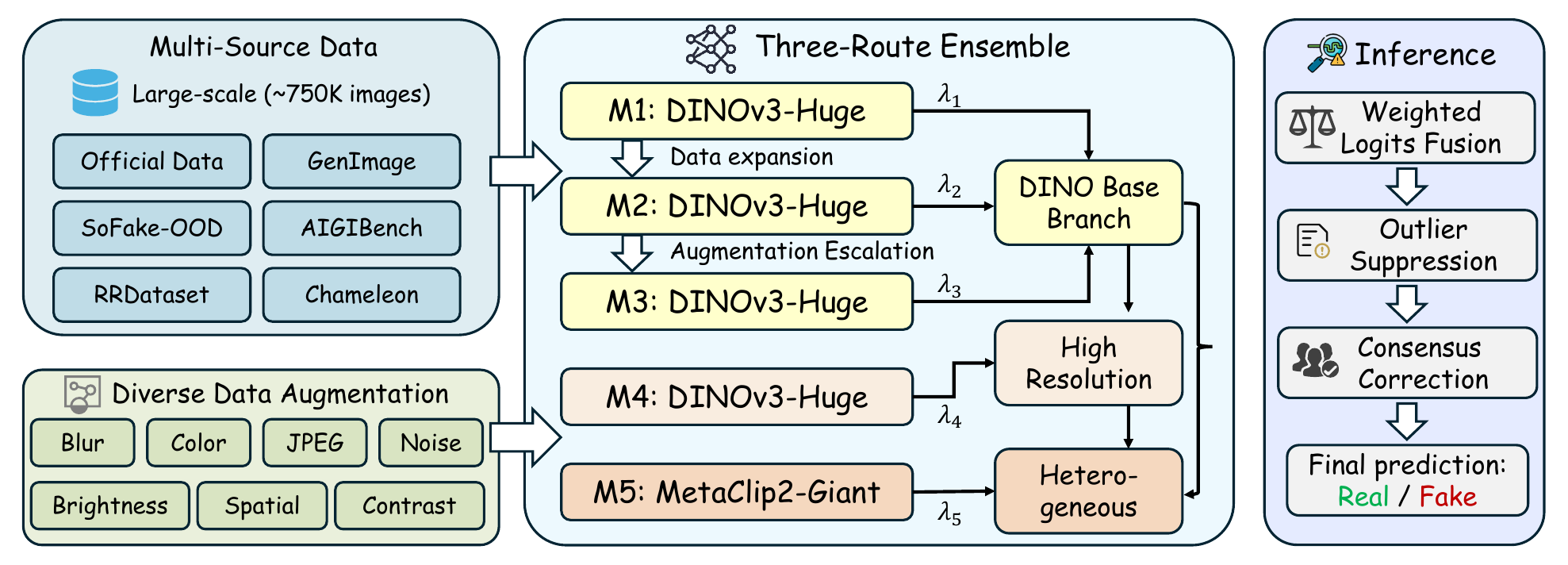}
    \caption{Overview of the proposed three-route framework for robust AIGC image detection. Route~A progressively constructs DINOv3-based detectors through data expansion and augmentation escalation, Route~B introduces a higher-resolution branch to capture fine-grained forensic cues, and Route~C adds a MetaCLIP2-based branch for backbone heterogeneity. Their outputs are fused in the logit space and further refined by a dual-gating mechanism.}
    \label{fig:overview}
\end{figure*}

\subsection{Overview}
\label{sec:overview}

We propose a three-route heterogeneous framework for robust AIGC image detection, as illustrated in Fig.~\ref{fig:overview}.
A single detector with a fixed backbone, resolution, and training configuration struggles to generalize across diverse data types and post-processing operations encountered in real-world scenarios.
Combining models that differ along multiple dimensions can mitigate this, as individual models tend to fail on different subsets of the data.
Following this principle, the framework introduces diversity along three dimensions: training data and augmentation, input resolution, and backbone architecture.
Specifically, Route~A progressively constructs DINOv3-Huge detectors (Models~1--3) through staged data expansion and augmentation escalation;
Route~B shares the same backbone but operates at a higher input resolution to preserve fine-grained forensic signals (Model~4);
and Route~C uses a different backbone, MetaCLIP2-Giant, to reduce correlation among branch predictions (Model~5).

All route outputs are aggregated via logit-space weighted fusion.
A dual-gating mechanism further refines the fused result by handling two failure patterns identified through systematic error analysis on the training set: occasional outlier predictions from the high-resolution branch, and cases where the Route~A majority overrides strong agreement between the two cross-route branches.

\subsection{Three-Route Heterogeneous Design}
\label{sec:three_routes}

\subsubsection{Route A: Progressive Training Evolution}
\label{sec:routeA}

A single training run tends to overfit to the particular data distribution and distortion profile seen during training.
Route~A mitigates this by constructing three detectors through continuation training on DINOv3-Huge: each subsequent stage is initialized from the previous checkpoint and trained with either more data or stronger augmentation.

Each model uses DINOv3-Huge with full-parameter fine-tuning, a lightweight two-layer MLP classification head, and cross-entropy loss.
The three stages are designed as follows:

\begin{itemize}
    \item \textbf{Stage~1 (Base).} The model is trained on the official competition data supplemented with SoFake-OOD~\cite{huang2025so}, which provides diverse generator coverage, and RRDataset~\cite{li2025bridging}, which contains samples with real-world degradation absent from the official set. This stage aims to build a solid foundation with broad data coverage.
    \item \textbf{Stage~2 (Data Expansion).} Training continues from Stage~1 with reduced learning rates. The training set is expanded to additionally include Chameleon~\cite{yan2025a}, GenImage~\cite{zhu2023genimage}, and AIGIBench~\cite{li2025artificial}, exposing the detector to a broader range of generator families and distribution shifts.
    \item \textbf{Stage~3 (Distortion Hardening).} Training continues from Stage~2 with stronger distortion augmentation: both the number and severity of randomly composed degradations (blur, JPEG compression, noise, color shifts, spatial transforms, etc.) are increased. This makes the detector more tolerant to realistic post-processing perturbations.
\end{itemize}

All three models share the same DINOv3-Huge backbone and are linked through continuation training, so they retain a common feature prior while developing different strengths: Stage~1 focuses on in-distribution accuracy, Stage~2 on cross-generator generalization, and Stage~3 on distortion robustness.
Including all three in the ensemble covers a wider range of conditions than any single checkpoint, effectively improving detection accuracy.

\subsubsection{Route B: Higher-Resolution Branch}
\label{sec:routeB}

Fine-grained forensic artifacts, such as subtle texture inconsistencies, boundary discontinuities, and compression traces, can be attenuated when images are downsampled to a standard resolution for model input.
Route~B uses the same DINOv3-Huge backbone as Route~A but increases the input resolution from $256\times256$ to $448\times448$, allowing the model to observe local details that are lost at lower resolutions.

In the final ensemble, Route~A operates at $256\times256$ and Route~B at $448\times448$.
The two scales are complementary: standard resolution is sufficient for most samples and keeps computation manageable, while higher resolution helps on samples whose forensic cues lie in fine local details that would otherwise be lost through downsampling.
This branch is also trained with cross-entropy loss.

\subsubsection{Route C: Backbone Heterogeneity}
\label{sec:routeC}

Routes~A and~B share the same DINOv3 backbone, so their feature representations, and consequently their error patterns, are inevitably correlated.
Reducing correlation among ensemble members generally improves collective performance, so Route~C introduces a structurally different backbone, MetaCLIP2-Giant.
DINOv3 is trained via self-supervised objectives on image patches, while MetaCLIP2 is trained through contrastive image-text alignment.
Because their pre-training objectives differ, the two backbones encode different visual priors, which helps diversify the predictions within the ensemble.

To avoid overwriting the pre-trained representations, we adopt a partial fine-tuning strategy: only the LayerNorm layers and the last two backbone blocks are updated, while the remaining parameters stay frozen.
This branch is optimized with Focal Loss:
\begin{equation}
\mathcal{L}_{\mathrm{Focal}} = -\sum_{c\in\{\text{real},\text{fake}\}} y_c (1-p_c)^{\gamma}\log p_c,
\end{equation}
where $\gamma{=}2$ is the focusing parameter.
Focal Loss down-weights well-classified examples and allocates more gradient to hard samples, which encourages this branch to focus on difficult cases.

\subsection{Logit-Space Ensemble}
\label{sec:ensemble}

We aggregate predictions from all five models in the logit space rather than averaging their output probabilities.
Given $M$ models with ensemble weights $\alpha_m$ ($\sum_m \alpha_m = 1$), the fused logit and final fake score are:
\begin{equation}
\label{eq:ensemble}
\begin{aligned}
\bar{\mathbf{z}}(x) &= \sum_{m=1}^{M} \alpha_m\,\mathbf{z}^{(m)}(x), \\
s(x) &= \mathrm{softmax}\!\left(\bar{\mathbf{z}}(x)\right)_{\text{fake}},
\end{aligned}
\end{equation}
where $\mathbf{z}^{(m)}(x)$ denotes the binary logits from model $m$.
We choose logit-space fusion over probability averaging because softmax compresses large logit differences into near-saturated probabilities.
Fusing after softmax can therefore obscure disagreements among branches, while fusing in the logit space retains the full dynamic range of each model's class evidence.

\paragraph{Weight assignment.}
The ensemble weights are assigned following a hierarchical scheme.
Within Route~A, Stage~1 serves as the base detector and receives the largest weight, while Stages~2 and~3 act as complementary refinements targeting cross-generator generalization and distortion robustness, respectively, and receive smaller weights.
Routes~A and~B are combined at a 7:3 ratio, and the resulting DINOv3-based prediction is further combined with Route~C at the same 7:3 ratio.

\paragraph{Test-time augmentation.}
For Models~3 and~4, we apply horizontal-flip TTA: the model processes both the original image and its flipped version, and their logits are averaged before entering the ensemble.
Horizontal flipping does not alter the authenticity of an image, so the detector should produce consistent predictions for both views; averaging the two reduces variance caused by orientation-dependent artifacts.

\subsection{Dual-Gating Robust Decision}
\label{sec:dual_gating}

Weighted logit fusion works well in most cases, but systematic error analysis on the training set reveals two recurring failure patterns.
We address them with a dual-gating mechanism based on directional logit evidence.

For model $i$, we define its directional evidence as
\begin{equation}
\label{eq:diff_logit}
d_i = \mathrm{logit}^{(i)}_{1} - \mathrm{logit}^{(i)}_{0},
\end{equation}
where a positive value indicates support for the fake class and a negative value for the real class.

\subsubsection{Gate-1: Outlier Suppression}
\label{sec:gate1}

The first failure pattern is that $M_4$ (the high-resolution branch) occasionally produces a prediction that contradicts all other branches.
Higher-resolution inputs preserve more local detail, but they also amplify local patterns, such as JPEG blocking artifacts or sensor noise, that can mislead the detector on individual samples.
$M_1$--$M_3$ and $M_5$, operating at lower resolutions or using a different backbone, are less affected by such local anomalies.

Gate-1 works as follows: if at least three models among $\{M_1, M_2, M_3, M_5\}$ agree on the prediction direction while $M_4$ disagrees, $M_4$ is excluded from the fusion and the weights are re-normalized over the remaining four models.
We apply this rule only to $M_4$ because training-set analysis shows it is the only branch that exhibits this isolated-disagreement pattern at a non-negligible rate.
The other branches share either the same resolution or backbone family and rarely contradict the majority alone.

\begin{table*}[!t]
\centering
\caption{Training and ensemble configurations for all five models. LR$_b$ and LR$_h$ denote the learning rates for the backbone and classification head, respectively.}
\label{tab:training_config}
\vspace{-2mm}
\begin{adjustbox}{max width=\textwidth}
\begin{tabular}{cccccccccccc}
\toprule
Model & Backbone & Fine-tune & Resolution & LR$_b$ & LR$_h$ & WD & Warmup & Loss & Epochs & Weight & TTA \\
\midrule
M1 & DINOv3-H & Full & $256\!\times\!256$ & $2\!\times\!10^{-5}$ & $5\!\times\!10^{-4}$ & 0.05/0.01 & 10\% & CE & 20 & 0.3675 & -- \\
M2 & DINOv3-H & Full & $256\!\times\!256$ & $5\!\times\!10^{-6}$ & $5\!\times\!10^{-5}$ & 0.05/0.01 & 10\% & CE & 20 & 0.0735 & -- \\
M3 & DINOv3-H & Full & $256\!\times\!256$ & $5\!\times\!10^{-6}$ & $5\!\times\!10^{-5}$ & 0.05/0.01 & 10\% & CE & 20 & 0.0490 & Flip \\
M4 & DINOv3-H & Full & $448\!\times\!448$ & $5\!\times\!10^{-5}$ & $1\!\times\!10^{-3}$ & 0.05/0.01 & 5\% & CE & 20 & 0.2100 & Flip \\
M5 & MetaCLIP2-G & Partial$^\ast$ & $378\!\times\!378$ & $1\!\times\!10^{-4}$ & $1\!\times\!10^{-4}$ & --- & --- & Focal & 20 & 0.3000 & -- \\
\bottomrule
\end{tabular}
\end{adjustbox}
\\[2pt]
\raggedright\footnotesize{$^\ast$ Only LayerNorm layers and the last two backbone blocks are fine-tuned.}
\end{table*}

\subsubsection{Gate-2: Cross-Route Consensus Correction}
\label{sec:gate2}

The second failure pattern occurs when the three Route~A models, which outnumber the other branches and share the same backbone, collectively dominate the fused logit, overriding the opposing evidence from $M_4$ (Route~B) and $M_5$ (Route~C).
Because $M_4$ and $M_5$ use different resolutions and backbones from Route~A, they are the two most structurally distinct branches in the ensemble.
When both produce high-confidence predictions in the same direction, this agreement between two structurally independent sources constitutes stronger evidence than their individual weights suggest.

Gate-2 works as follows: if $M_4$ and $M_5$ agree on the prediction direction with high confidence ($|d_4| \ge \tau_1$ and $|d_5| \ge \tau_2$), but the fused logit points in the opposite direction, we apply a corrective shift of $\delta$ to the fused logit in the direction indicated by $M_4$ and $M_5$.

In summary, Gate-1 suppresses spurious predictions from Route~B, and Gate-2 corrects cases where the Route~A majority overrides cross-route evidence.
The final fake probability is computed by softmax as in Eq.~\ref{eq:ensemble}.

\section{Experiments}
\label{sec:experiments}

\subsection{Experimental Setup}
\label{sec:setup}

\paragraph{Datasets.}
We use different data configurations for each model to maximize diversity.
M1 is trained on the official competition data supplemented with SoFake-OOD~\cite{huang2025so}, which covers 30 generation and manipulation methods across 12 diverse categories providing broad generator-type diversity, and RRDataset~\cite{li2025bridging}, which contains samples degraded by real-world social media propagation absent from the official set.
M2 continues from M1 and further includes Chameleon~\cite{yan2025a}, GenImage~\cite{zhu2023genimage}, and AIGIBench~\cite{li2025artificial} to expose the detector to a broader range of generator families and distribution shifts.
M3 shares the same data as M2 but applies stronger augmentation.
M4 and M5 are trained on the official dataset only.
For evaluation, we adopt three categories of benchmarks following the protocols of MIRROR~\cite{liu2026mirror} and REM~\cite{liu2025beyond}:
(i)~\textit{standard benchmarks}---GenImage$^\dagger$~\cite{zhu2023genimage}, AIGCDetect~\cite{zhong2023patchcraft}, DRCT-2M~\cite{chen2024drct}, Synthbuster~\cite{bammey2023synthbuster}, and EvalGEN~\cite{chen2025dual}---which test cross-generator generalization;
(ii)~\textit{in-the-wild benchmarks}---Chameleon$^\dagger$~\cite{yan2025a}, RRDataset$^\ddagger$~\cite{li2025bridging}, AIGIBench$^\dagger$~\cite{li2025artificial}, BFree-Online~\cite{guillaro2025bias}, SynthWildx~\cite{cozzolino2024raising}, and WildRF~\cite{cavia2024real}---which contain images degraded by real-world social media propagation;
and (iii)~\textit{chain degradation} on RealChain~\cite{liu2025beyond}, which simulates multi-hop cross-platform transmission and user post-processing.
Benchmarks marked with $^\dagger$ share data sources with M2/M3's training set, and $^\ddagger$ with M1's training set; all remaining benchmarks are entirely unseen during training.

\paragraph{Implementation details.}
All models are trained on 8$\times$H800 GPUs with DDP.
The detailed per-model training configurations are summarized in Table~\ref{tab:training_config}.
All DINOv3-based models (M1--M4) use AdamW ($\beta{=}(0.9,0.999)$) with a warmup-cosine learning rate scheduler.
Route~C (M5) uses a learning rate of $1\times10^{-4}$ with Focal Loss.
For Routes~A and~B, the classifier is a two-layer MLP: $\mathrm{Linear}(1280\!\rightarrow\!256)\!\rightarrow\!\mathrm{ReLU}\!\rightarrow\!\mathrm{Dropout}(0.1)\!\rightarrow\!\mathrm{Linear}(256\!\rightarrow\!2)$.
For Route~C, a deeper MLP is used: $1664\!\rightarrow\!2048\!\rightarrow\!512\!\rightarrow\!256\!\rightarrow\!2$.
All models apply the official competition distortion augmentation pipeline, which randomly composes degradations sampled from seven groups (blur, color shift, JPEG compression, noise, brightness change, spatial distortion, and contrast change), with severity levels drawn from a Gaussian-weighted distribution.
M1 and M2 use moderate settings (\texttt{max\_distortions}{=}3, \texttt{num\_levels}{=}3); M3 increases both to 5 for stronger distortion hardening; M4 uses \texttt{max\_distortions}{=}3 with \texttt{num\_levels}{=}5, applying fewer operations at higher severity; and M5 applies moderate augmentation at a reduced probability (\texttt{aug\_prob}{=}0.2).
The Gate-2 thresholds are set to $\tau_1{=}8$, $\tau_2{=}3$, and the correction magnitude to $\delta{=}2.5$.

\paragraph{Evaluation metrics and comparative methods.}
We report three metrics: Balanced Accuracy (B.Acc), JPEG Robustness (J.Rob) under QF=90 compression, and Resize Robustness (R.Rob) under resize factor 0.9.
We compare against a broad set of state-of-the-art detectors spanning artifact-based methods (NPR~\cite{tan2024rethinking}, UnivFD~\cite{ojha2023towards}, FatFormer~\cite{liu2024forgeryaware}, SAFE~\cite{li2025improving}, C2P-CLIP~\cite{tan2025c2p}, AIDE~\cite{yan2025a}), dataset-alignment-based methods (DRCT~\cite{chen2024drct}, Aligned~\cite{rajan2024aligned}, B-Free~\cite{guillaro2025bias}, DDA~\cite{chen2025dual}), and VFM-based methods (MIRROR~\cite{liu2026mirror}, REM~\cite{liu2025beyond}).
We adopt the same evaluation datasets and protocols as MIRROR~\cite{liu2026mirror} and REM~\cite{liu2025beyond}, and directly cite the baseline results reported therein to ensure a fair comparison.

\begin{table*}[t!]
\centering
\caption{Performance comparison on \textbf{Standard Benchmarks}. We report Balanced Accuracy (B.Acc), JPEG Robustness (J.Rob), and Resize Robustness (R.Rob). The best and second-best results in the summary columns are highlighted in \textbf{bold} and \underline{underline}, respectively. $^\dagger$Benchmark shares data sources with the training set of M2/M3. Baseline results are cited from MIRROR~\cite{liu2026mirror}. We exclude UnivFakeDetect due to its outdated generator coverage.}
\vspace{-2mm}
\fontsize{9pt}{11pt}\selectfont
\setlength{\tabcolsep}{5pt}
\renewcommand{\arraystretch}{1}
\begin{adjustbox}{max width=\textwidth}
\begin{tabular}{l | c c c | c c c | c c c | c c c | c c c | c c c}
\toprule
\multirow{2}{*}{\textbf{Method}} &
\multicolumn{3}{c|}{\textbf{GenImage$^\dagger$}} &
\multicolumn{3}{c|}{\textbf{AIGCDetect}} &
\multicolumn{3}{c|}{\textbf{DRCT-2M}} &
\multicolumn{3}{c|}{\textbf{Synthbuster}} &
\multicolumn{3}{c|}{\textbf{EvalGEN}} &
\multicolumn{3}{c}{\textbf{Average}} \\
\cmidrule(lr){2-4} \cmidrule(lr){5-7} \cmidrule(lr){8-10} \cmidrule(lr){11-13} \cmidrule(lr){14-16} \cmidrule(lr){17-19}
 & B.Acc & J.Rob & R.Rob & B.Acc & J.Rob & R.Rob & B.Acc & J.Rob & R.Rob & B.Acc & J.Rob & R.Rob & B.Acc & J.Rob & R.Rob & B.Acc & J.Rob & R.Rob \\
\midrule
NPR~\cite{tan2024rethinking}        & 73.7 & 73.7 & 46.6 & 63.9 & 64.0 & 48.2 & 59.5 & 58.3 & 37.4 & 64.6 & 64.0 & 34.4 & 66.1 & 65.7 & 64.2 & 65.6 & 65.1 & 46.2 \\
UnivFD~\cite{ojha2023towards}       & 62.5 & 53.6 & 50.8 & 56.5 & 50.1 & 48.7 & 69.5 & 57.3 & 52.5 & 65.8 & 59.2 & 52.3 & 73.8 & 61.3 & 56.6 & 65.6 & 56.3 & 52.2 \\
FatFormer~\cite{liu2024forgeryaware}     & 71.5 & 58.4 & 68.4 & 82.1 & 71.5 & 80.0 & 53.9 & 52.1 & 52.4 & 69.2 & 57.1 & 65.6 & 56.7 & 53.4 & 57.5 & 66.7 & 58.5 & 64.8 \\
SAFE~\cite{li2025improving}         & 47.7 & 64.4 & 49.4 & 48.4 & 57.8 & 49.4 & 50.3 & 62.4 & 49.2 & 51.6 & 53.8 & 49.9 & 50.2 & 55.4 & 49.1 & 49.6 & 58.8 & 49.4 \\
C2P-CLIP~\cite{tan2025c2p}          & 71.1 & 55.6 & 67.8 & 80.0 & 65.7 & 77.6 & 54.4 & 54.0 & 55.7 & 69.4 & 57.3 & 66.4 & 56.8 & 52.1 & 65.8 & 66.3 & 56.9 & 66.7 \\
AIDE~\cite{yan2025a}           & 88.6 & 55.3 & 76.3 & 84.0 & 53.4 & 77.0 & 59.2 & 54.6 & 72.5 & 75.4 & 76.8 & 62.5 & 59.5 & 58.8 & 76.4 & 73.3 & 59.8 & 72.9 \\
DRCT~\cite{chen2024drct}            & 78.8 & 78.5 & 80.3 & 67.1 & 67.1 & 67.5 & 97.0 & 93.0 & 98.4 & 81.0 & 80.6 & 83.3 & 81.9 & 85.5 & 82.4 & 81.2 & 80.9 & 82.4 \\
Aligned~\cite{rajan2024aligned}     & 57.5 & 55.9 & 57.0 & 55.7 & 54.5 & 55.3 & 54.9 & 54.7 & 54.3 & 54.5 & 53.0 & 53.8 & 65.8 & 65.2 & 67.5 & 57.7 & 56.7 & 57.6 \\
B-Free~\cite{guillaro2025bias}      & 89.6 & 90.2 & 87.9 & 84.7 & 85.0 & 82.7 & 99.2 & 99.0 & 98.2 & 95.7 & 95.9 & 94.8 & 94.6 & 93.7 & 92.5 & 92.8 & 92.8 & 91.2 \\
DDA~\cite{chen2025dual}             & 88.9 & 88.7 & 88.1 & 81.5 & 81.8 & 79.1 & 97.0 & 97.7 & 95.6 & 96.5 & 95.6 & 94.6 & 96.6 & 90.1 & 93.9 & 92.1 & 90.8 & 90.3 \\
MIRROR~\cite{liu2026mirror}                     & 94.2 & 96.7 & 94.9 & 91.7 & 91.5 & 91.5 & 93.0 & 90.4 & 93.4 & 98.1 & 97.5 & 97.6 & 99.0 & 98.5 & 98.6 & \underline{95.2} & \underline{94.9} & \underline{95.2} \\
\midrule
\rowcolor{blue!5}
\textbf{HEDGE}                     & 99.8 & 99.7 & 99.9 & 99.5 & 99.1 & 99.6 & 95.0 & 93.8 & 95.4 & 97.3 & 96.8 & 97.2 & 99.9 & 99.8 & 99.9 & \textbf{98.3}{\scriptsize\textcolor{red}{$\uparrow$3.1}} & \textbf{97.8}{\scriptsize\textcolor{red}{$\uparrow$2.9}} & \textbf{98.4}{\scriptsize\textcolor{red}{$\uparrow$3.2}} \\
\bottomrule
\end{tabular}
\end{adjustbox}
\label{tab:standard_benchmarks}
\end{table*}

\begin{table*}[t!]
\centering
\caption{Performance comparison on \textbf{In-the-wild Benchmarks}. These datasets have undergone low-level degradation caused by network propagation. The best and second-best results in the summary columns are highlighted in \textbf{bold} and \underline{underline}, respectively. $^\dagger$Benchmark shares data sources with M2/M3's training set; $^\ddagger$with M1's training set. Baseline results are cited from MIRROR~\cite{liu2026mirror}. We exclude CO-SPY-Bench as the dataset is not publicly released.}
\vspace{-2mm}
\fontsize{9pt}{11pt}\selectfont
\setlength{\tabcolsep}{5pt}
\renewcommand{\arraystretch}{1}
\begin{adjustbox}{max width=\textwidth}
\begin{tabular}{l c | c | c c | c | c c c | c c c | c}
\toprule
\multirow{2}{*}{\textbf{Method}} &
\multirow{2}{*}{\textbf{\makecell{Cham-\\eleon$^\dagger$}}} &
\multirow{2}{*}{\textbf{\makecell{RR-\\Dataset$^\ddagger$}}} &
\multicolumn{2}{c|}{\textbf{AIGIBench$^\dagger$}} &
\multirow{2}{*}{\textbf{\makecell{BFree-\\Online}}} &
\multicolumn{3}{c|}{\textbf{SynthWildx}} &
\multicolumn{3}{c|}{\textbf{WildRF}} &
\multirow{2}{*}{\textbf{\makecell{Avg\\B.Acc}}}
\\
\cmidrule(lr){4-5} \cmidrule(lr){7-9} \cmidrule(lr){10-12}

 & & & SocRF & ComAI & & DALLE3 & Firefly & Midj. & FB & Reddit & Twitter & \\
\midrule
NPR~\cite{tan2024rethinking}      & 55.2 & 48.3 & 55.8 & 52.0 & 40.5 & 62.9 & 53.8 & 63.0 & 53.8 & 53.7 & 57.7 & 54.2 \\
UnivFD~\cite{ojha2023towards}     & 39.5 & 51.1 & 51.6 & 45.4 & 57.2 & 52.5 & 52.4 & 50.7 & 54.1 & 55.1 & 66.5 & 52.4 \\
FatFormer~\cite{liu2024forgeryaware}   & 57.8 & 50.4 & 55.7 & 50.4 & 32.7 & 52.6 & 56.6 & 50.0 & 52.5 & 65.4 & 39.7 & 51.3 \\
SAFE~\cite{li2025improving}       & 56.8 & 49.3 & 49.2 & 49.3 & 32.6 & 48.8 & 46.6 & 48.7 & 49.7 & 49.2 & 34.7 & 46.8 \\
C2P-CLIP~\cite{tan2025c2p}        & 57.6 & 50.0 & 58.1 & 50.4 & 32.7 & 49.6 & 57.9 & 49.6 & 51.9 & 67.6 & 40.4 & 51.4 \\
AIDE~\cite{yan2025a}         & 65.7 & 57.6 & 59.2 & 62.2 & 52.1 & 66.4 & 48.2 & 66.4 & 61.6 & 66.3 & 52.5 & 59.8 \\
DRCT~\cite{chen2024drct}          & 79.8 & 58.2 & 71.3 & 84.6 & 77.1 & 85.9 & 58.9 & 90.5 & 90.3 & 66.8 & 79.6 & 76.6 \\
Aligned~\cite{rajan2024aligned}   & 61.3 & 47.7 & 51.0 & 60.2 & 38.1 & 49.6 & 53.9 & 52.4 & 48.4 & 54.0 & 40.6 & 50.7 \\
B-Free~\cite{guillaro2025bias}    & 78.3 & 69.5 & 84.9 & 79.7 & 87.1 & 96.1 & 92.3 & 95.3 & 95.6 & 85.5 & 96.7 & 87.4 \\
DDA~\cite{chen2025dual}           & 83.5 & 70.3 & 79.9 & 88.9 & 81.2 & 91.1 & 84.7 & 91.6 & 85.3 & 82.5 & 89.3 & 84.4 \\
MIRROR~\cite{liu2026mirror}                  & 90.7 & 78.9 & 87.6 & 93.4 & 83.0 & 95.9 & 88.4 & 94.9 & 97.1 & 96.6 & 96.4 & \underline{91.2} \\
\midrule
\rowcolor{blue!5}
\textbf{HEDGE}                   & 99.9 & 99.9 & 98.3 & 99.9 & 82.1 & 97.9 & 91.8 & 97.7 & 99.3 & 98.8 & 99.2 & \textbf{96.8}{\scriptsize\textcolor{red}{$\uparrow$5.6}} \\
\bottomrule
\end{tabular}
\end{adjustbox}
\label{tab:wild}
\vspace{-3mm}
\end{table*}

\begin{figure*}[t]
    \centering
    \includegraphics[width=\linewidth]{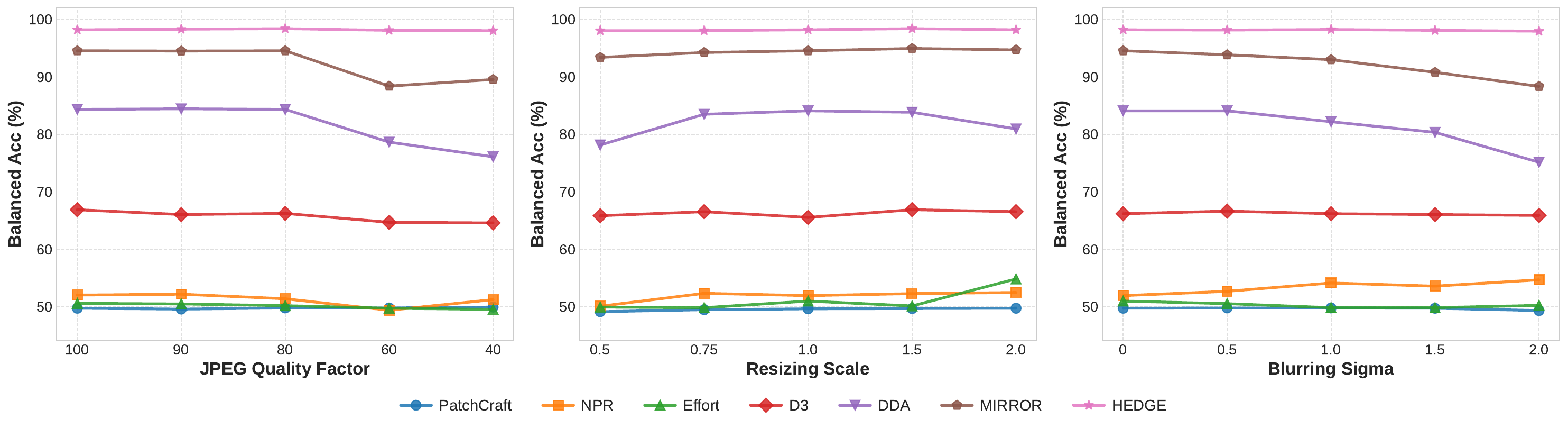}
    \caption{Robustness evaluation under common image perturbations on HiRes-50K (1,000 real + 1,000 fake, unseen during training). We report B.Acc under JPEG compression (left), spatial resizing (middle), and Gaussian blurring (right) at varying intensities. HEDGE maintains near-constant performance across all conditions.}
    \label{fig:robustness}
\end{figure*}

\subsection{Main Results}
\label{sec:main_results}

\paragraph{Standard benchmarks.}
Table~\ref{tab:standard_benchmarks} compares HEDGE against state-of-the-art methods on five standard benchmarks.
HEDGE achieves an average B.Acc of 98.3\%, outperforming the second-best method MIRROR~\cite{liu2026mirror} by +3.1 points.
More importantly, the gap between B.Acc (98.3\%), J.Rob (97.8\%), and R.Rob (98.4\%) is minimal, indicating that the detection performance is not inflated by fragile cues that break under compression or resizing.
Among the four \emph{unseen} benchmarks, HEDGE achieves near-perfect scores on AIGCDetect (99.5\%) and EvalGEN (99.9\%), both of which span diverse generator architectures from GANs to autoregressive models.
On DRCT-2M, HEDGE outperforms MIRROR (95.0\% vs.\ 93.0\% B.Acc); the smaller margin is expected, as DRCT-2M focuses exclusively on diffusion-based generators where both VFM-based methods already achieve strong performance.
On Synthbuster, HEDGE (97.3\%) is slightly below MIRROR (98.1\%), though the gap is marginal and both methods substantially outperform all other baselines.

\begin{table}[t!]
\centering
\caption{Comparison on the \textit{RealChain} under \textbf{Chain Degradations}. The best and second-best results for Balanced Accuracy (B.Acc) are highlighted in \textbf{bold} and \underline{underline}, respectively. Results of baseline methods are taken from~\cite{liu2025beyond}.}
\vspace{-2mm}
\begin{adjustbox}{max width=\linewidth}
\begin{tabular}{l|ccc}
\toprule
\multirow{2}{*}{\textbf{Method}} &
\multicolumn{3}{c}{\textbf{Chain Degradations}} \\
\cmidrule(lr){2-4}
& R.Acc & F.Acc & B.Acc \\
\midrule
NPR~\cite{tan2024rethinking}               & 73.5 & 37.9 & 55.7 \\
UnivFD~\cite{ojha2023towards}            & 95.8 &  6.9 & 51.3 \\
FatFormer~\cite{liu2024forgeryaware}         & 98.3 &  4.1 & 51.2 \\
SAFE~\cite{li2025improving}              & 99.3 &  0.3 & 49.8 \\
C2P-CLIP~\cite{tan2025c2p}          & 98.3 &  4.3 & 51.3 \\
AIDE~\cite{yan2025a}              & 98.8 &  1.3 & 50.0 \\
DRCT~\cite{chen2024drct}              & 92.0 & 18.9 & 55.4 \\
Aligned~\cite{rajan2024aligned}           & 99.8 & 16.4 & 58.0 \\
DDA~\cite{chen2025dual}               & 79.3 & 52.4 & 65.8 \\
REM~\cite{liu2025beyond}           & 85.3 & 83.0 & \underline{84.2} \\
\midrule
\rowcolor{blue!5}
\textbf{HEDGE}    & 98.7 & 87.8 & \textbf{93.2}{\scriptsize\textcolor{red}{$\uparrow$9.0}} \\
\bottomrule
\end{tabular}
\end{adjustbox}
\label{tab:chain_degradations}
\vspace{-3mm}
\end{table}

\paragraph{In-the-wild benchmarks.}
Table~\ref{tab:wild} presents results under real-world deployment conditions, where images are produced by more advanced generators and further undergo uncontrolled degradation from social media propagation, making detection considerably harder than on standard benchmarks.
HEDGE achieves an average B.Acc of 96.8\%, surpassing MIRROR by +5.6 points.
Among the \emph{unseen} benchmarks, the advantage is especially pronounced on WildRF, where HEDGE achieves 99.3\%, 98.8\%, and 99.2\% on Facebook, Reddit, and Twitter subsets respectively, demonstrating consistent robustness across different platform-specific degradation pipelines.
On SynthWildx, HEDGE achieves 97.9\% on DALLE3 and 97.7\% on Midjourney, outperforming all baselines including the alignment-based B-Free (96.1\%/95.3\%); the lower score on Firefly (91.8\%) suggests that certain proprietary generation pipelines with distinctive post-processing still pose challenges.
The main weakness is BFree-Online (82.1\%), which contains a high proportion of local inpainting and self-conditioned generation samples---a fundamentally different task from whole-image generation that our framework targets.
This limitation is shared across most baselines (only B-Free reaches 87.1\%, as BFree-Online is the evaluation set introduced alongside B-Free and its training incorporates inpainting-aware data), suggesting that local manipulation detection requires dedicated modeling beyond what global classification provides.

\paragraph{Chain degradation.}
Table~\ref{tab:chain_degradations} reports results on the chain-degraded subset of RealChain~\cite{liu2025beyond}, which simulates realistic multi-stage degradation chains including cross-platform compression, re-encoding, and user post-processing.
Since the original no-degradation subset is not publicly available, we evaluate all methods on this degraded subset only, making it a strict test of robustness under compound distortions.
This scenario exposes a critical failure mode of existing detectors: most baselines suffer catastrophic drops in Fake Accuracy (near 0\%) while maintaining high Real Accuracy ($>$90\%), revealing a strong bias toward predicting ``real'' under heavy degradation.
The compound distortions effectively wash out the low-level forensic signals that single-model detectors rely on, causing them to default to the real class.
In contrast, HEDGE achieves a B.Acc of 93.2\% (+9.0 over REM), maintaining a Fake Accuracy of 87.8\% alongside 98.7\% Real Accuracy---a much more balanced prediction profile.
This result is consistent with our design motivation: the combination of distortion-hardened training (Route~A Stage~3), multi-scale features (Route~B), and backbone diversity (Route~C) provides complementary forensic evidence that remains informative even when individual signal types are severely degraded.
Notably, REM~\cite{liu2025beyond} also leverages DINOv3 features but achieves a substantially lower B.Acc (84.2\%), suggesting that a single backbone with a single detection paradigm may face inherent limitations under extreme compound degradation.

\begin{table}[t!]
\centering
\caption{Ablation study on the competition public test set (1,250 clean + 1,250 distorted samples). The best result in each column is highlighted in \textbf{bold}.}
\vspace{-2mm}
\begin{adjustbox}{max width=\textwidth}
\begin{tabular}{l|cc|cc}
\toprule
\multirow{2}{*}{\textbf{Configuration}} & \multicolumn{2}{c|}{\textbf{Clean}} & \multicolumn{2}{c}{\textbf{Robust}} \\
\cmidrule(lr){2-3} \cmidrule(lr){4-5}
 & AUC & F1 & AUC & F1 \\
\midrule
\multicolumn{5}{l}{\textit{(a) Route A: Progressive Training}} \\
\midrule
M1 & 98.59 & 92.62 & 89.16 & 77.77 \\
M2 & 98.11 & 93.29 & 87.97 & 77.90 \\
M3  & 98.73 & 93.78 & 89.67 & 79.24 \\
Route A & 98.80 & 93.92 & 89.95 & 80.03 \\
\midrule
\multicolumn{5}{l}{\textit{(b) Multi-Route Integration}} \\
\midrule
+ Route B & 98.94 & 93.81 & 90.41 & 79.95 \\
+ Route C & 98.95 & 93.84 & 90.55 & 80.61 \\
Full (A+B+C) & 98.98 & 93.91 & 90.87 & 80.52 \\
\midrule
\multicolumn{5}{l}{\textit{(c) Dual-Gating Mechanism}} \\
\midrule
+ Gate-1 only & \textbf{98.99} & 93.94 & 90.88 & 80.58 \\
+ Gate-2 only & 98.98 & 94.00 & 90.83 & \textbf{80.89} \\
\midrule
\multicolumn{5}{l}{\textit{(d) Fusion Strategy}} \\
\midrule
Majority voting & -- & 93.31 & -- & 79.83 \\
Prob. averaging & 98.84 & 93.14 & 89.83 & 79.83 \\
Equal-weight logit & 98.95 & 92.51 & 90.49 & 78.83 \\
\rowcolor{blue!5}
HEDGE & \textbf{98.99} & \textbf{94.01} & \textbf{90.91} & \textbf{80.89} \\
\bottomrule
\end{tabular}
\end{adjustbox}
\label{tab:ablation}
\vspace{-3mm}
\end{table}

\subsection{Robustness Analysis}
\label{sec:robustness}

To further evaluate robustness under controlled perturbations, we conduct stress tests on a held-out subset of HiRes-50K~\cite{mu2026no}, a high-resolution benchmark containing over 50K images (up to 64 megapixels) collected from online AIGC communities with real-fake pairs aligned in resolution and JPEG compression level, entirely unseen during training.
We randomly sample 1,000 real and 1,000 fake images and apply three common post-processing operations at varying intensities: JPEG compression (QF from 100 to 40), spatial resizing (scale from $0.5\times$ to $2.0\times$), and Gaussian blurring ($\sigma$ from 0 to 2.0).

As shown in Fig.~\ref{fig:robustness}, HEDGE exhibits remarkable stability across all perturbation types and intensities, maintaining a B.Acc above 97.9\% even under the most aggressive settings (JPEG QF=40, resize $0.5\times$, blur $\sigma$=2.0).
In contrast, MIRROR suffers noticeable degradation under strong JPEG compression (94.6\% $\rightarrow$ 89.6\% at QF=40) and heavy blurring (94.6\% $\rightarrow$ 88.4\% at $\sigma$=2.0), while DDA drops more sharply (84.4\% $\rightarrow$ 76.1\% at QF=40; 84.1\% $\rightarrow$ 75.2\% at $\sigma$=2.0).
Other artifact-based methods (PatchCraft~\cite{zhong2023patchcraft}, NPR~\cite{tan2024rethinking}, Effort~\cite{yan2024orthogonal}) remain near chance level regardless of perturbation intensity.

The near-flat response curve of HEDGE across all conditions suggests that the heterogeneous ensemble captures forensic representations that are largely invariant to common post-processing, rather than relying on fragile low-level cues that are easily disrupted by compression, interpolation, or smoothing.

\subsection{Ablation Study}
\label{sec:ablation}

We conduct ablation experiments on the competition public test set to analyze each component. Results are in Table~\ref{tab:ablation}.

\paragraph{Route~A progressive training (a).}
Each successive stage improves Robust F1 (77.77$\rightarrow$77.90$\rightarrow$79.24), and the Route~A ensemble further raises it to 80.03, confirming that continuation training produces complementary detectors.

\paragraph{Multi-route integration (b).}
Adding Route~B and Route~C each improves AUC consistently (89.95$\rightarrow$90.41/90.55), and the full ensemble achieves the highest AUC (90.87 robust).
Route~C contributes the largest Robust F1 gain (+0.58), validating that backbone diversity is particularly beneficial under distortion.

\paragraph{Dual-gating (c).}
Compared to the full ensemble without gating in~(b), Gate-1 provides modest improvements in AUC, while Gate-2 yields more substantial F1 gains (93.91$\rightarrow$94.00 clean, 80.52$\rightarrow$80.89 robust), confirming that cross-route consensus correction effectively recovers predictions that raw fusion misclassifies.

\paragraph{Fusion strategy (d).}
Weighted logit fusion (HEDGE) consistently outperforms probability averaging and majority voting, confirming that fusing in the logit space with hierarchical weights better preserves inter-branch disagreement signals and yields more accurate decisions.

\subsection{Feature Space Visualization}
\label{sec:visualization}

\begin{figure}[t]
    \centering
    \includegraphics[width=\linewidth]{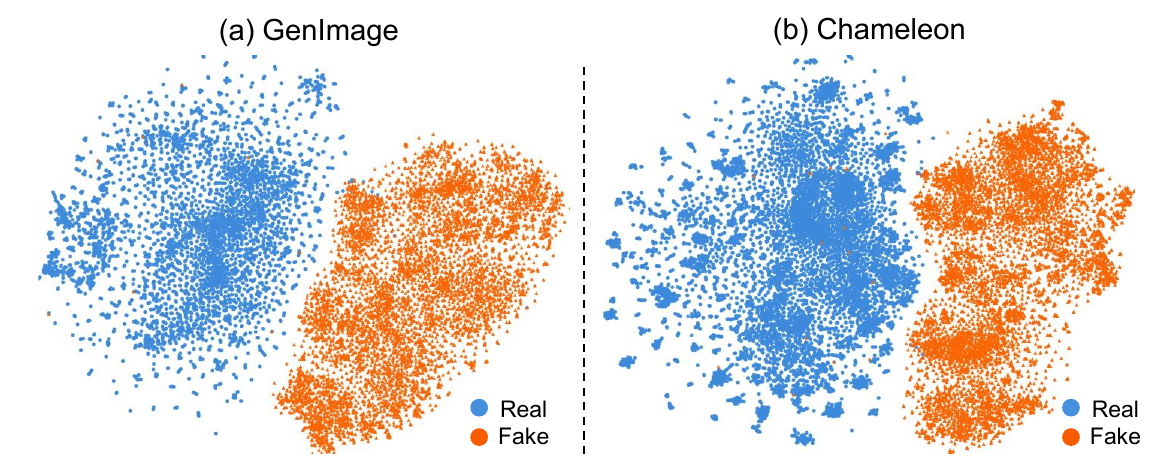}
    \caption{t-SNE visualization of M3 (DINOv3-Huge) CLS token features on (a) GenImage and (b) Chameleon.}
    \label{fig:tsne}
\end{figure}

We visualize the CLS token features of M3 (DINOv3-Huge) using t-SNE on GenImage and Chameleon.
As shown in Fig.~\ref{fig:tsne}, real and fake images form nearly fully separable clusters on both benchmarks, confirming that the fine-tuned DINOv3 features encode strong discriminative representations for AIGC detection.

\section{Conclusion}
\label{sec:conclusion}

We present HEDGE, a heterogeneous ensemble framework for robust AI-generated image detection in the wild.
HEDGE introduces structured diversity along three complementary axes: progressive DINOv3-based training with data expansion and augmentation escalation (Route~A), higher-resolution feature extraction (Route~B), and backbone heterogeneity via MetaCLIP2 (Route~C).
A logit-space weighted fusion with a lightweight dual-gating mechanism further refines predictions by suppressing branch-level outliers and correcting majority-dominated errors.
Extensive experiments on standard, in-the-wild, and chain-degradation benchmarks demonstrate that HEDGE achieves state-of-the-art detection accuracy and robustness, substantially outperforming existing methods under diverse post-processing conditions.

\clearpage
{
    \small
    \bibliographystyle{ieeenat_fullname}
    \bibliography{main}
}


\end{document}